\documentclass[10pt, conference, compsocconf]{IEEEtran}

\usepackage{cite}
\usepackage{amsmath,amssymb,amsfonts}
\usepackage{algorithmic}
\usepackage{graphicx}
\usepackage{textcomp}
\usepackage{xcolor}
\usepackage{multirow}
\usepackage{bm}
\usepackage{subfigure}
\usepackage{booktabs}
\def\BibTeX{{\rm B\kern-.05em{\sc i\kern-.025em b}\kern-.08em
    T\kern-.1667em\lower.7ex\hbox{E}\kern-.125emX}}
\begin{document}

\title{Edge-Level Explanations for Graph Neural Networks by Extending \\Explainability Methods for Convolutional Neural Networks}

\author{\IEEEauthorblockN{Tetsu Kasanishi\IEEEauthorrefmark{1}, Xueting Wang\IEEEauthorrefmark{1}, and Toshihiko Yamasaki\IEEEauthorrefmark{1}}
\IEEEauthorblockA{\IEEEauthorrefmark{1}Department of Information and Communication Engineering\\
The University of Tokyo, Tokyo, Japan\\
Email: kasanishi-tetsu9923@g.ecc.u-tokyo.ac.jp, xt\_wang@hal.t.u-tokyo.ac.jp, yamasaki@cvm.t.u-tokyo.ac.jp\\
}}

\maketitle

\begin{abstract}
Graph Neural Networks (GNNs) are deep learning models that take graph data as inputs, and they are applied to various tasks such as traffic prediction and molecular property prediction. 
However, owing to the complexity of the GNNs, it has been difficult to analyze which parts of inputs affect the GNN model's outputs.
In this study, we extend explainability methods for Convolutional Neural Networks (CNNs), such as Local Interpretable Model-Agnostic Explanations (LIME), Gradient-Based Saliency Maps, and Gradient-Weighted Class Activation Mapping (Grad-CAM) to GNNs, and predict which edges in the input graphs are important for GNN decisions.
The experimental results indicate that the LIME-based approach is the most efficient explainability method for multiple tasks in the real-world situation, outperforming even the state-of-the-art method in GNN explainability. 
\end{abstract}

\begin{IEEEkeywords}
Graph Neural Networks, Explainability, LIME, Saliency Maps, Grad-CAM
\end{IEEEkeywords}

\section{INTRODUCTION}
In recent years, there has been considerable number of studies with regard to explaining the decisions of deep learning models.
While deep learning models have been used to improve the accuracy of various tasks, they encounter a challenge: it is difficult to understand the basis of their decisions.
This makes it difficult to use deep learning models for tasks that require explanations, such as medical image processing.
Explanations are also helpful in understanding the model's behavior.
For these reasons, research on understanding the rationale for decisions of deep learning models has been widely conducted.

GNNs are deep learning models that take graph data as inputs.
In many real-world situations, data are represented in the form of graphs.
For example, molecular structures can be represented as graphs where nodes are atoms and edges are chemical bonds.
Therefore, GNNs are becoming powerful tools that can be applied to a variety of tasks such as drug discovery.
However, similar to other deep learning models, GNNs cannot present the reasoning behind their decisions.

In this study, we extend several explainability methods for CNNs to GNNs to calculate the importance of the edges for the models' outputs.
Reference \cite{survey} states that graph convolution in GNNs is generalized from 2-D convolution in CNNs because both take the weighted sum of information from neighborhood nodes/pixels.
This similarity between GNNs and CNNs makes it reasonable to apply techniques used for CNNs to GNNs.
Thus, we investigate LIME~\cite{LIME}, Gradient-Based Saliency Maps~\cite{saliency-maps}, and Grad-CAM~\cite{grad-cam}. These are frequently used in computer vision tasks. 
Although the techniques specified in this study are not novel, our contribution is that we extend off-the-shelf explainability methods in computer vision to GNNs and experimentally show that LIME is the best approach. Furthermore, we found LIME to be superior to a state-of-the-art method~\cite{GNNExplainer}.

\section{RELATED WORK}\label{sec:related}

\subsection{Formulation of GNNs}
Although a variety of GNN methods have been proposed, most of them can be expressed in the framework of message passing~\cite{MP} as follows.
Let $G$ be an input graph of GNNs, $\bm{h}_v^t$ be the feature vector of the node $v$ in $G$ at the $t$th message passing phase, and $N(v)$ be the set of nodes adjacent to the node $v$.
The message passing operation is expressed by the following equation:
\begin{eqnarray}
\bm{h}_{v}^{t+1} &=& U_{t}\left(\bm{h}_{v}^{t}, \sum_{w \in N(v)} M_{t}\left(\bm{h}_{v}^{t}, \bm{h}_{w}^{t}\right)\right).
\end{eqnarray}
Here, $M_{t}$ and $U_{t}$ are functions defined for different methods, where $M_{t}$ collects information from neighboring nodes, and $U_{t}$ updates the feature vector of each node based on the neighboring information.
By performing these message passing operations $T$ times, a higher-order feature vector $\bm{h}_{v}^{T}$ of the node $v$ can be obtained.

For a graph classification task, the feature vector for the entire graph is then calculated from each node's feature vector by taking its summation or mean, for example.
By performing the above-mentioned operations, feature vectors for each node or each graph can be obtained.
Finally, these feature vectors are fed into fully connected layers.

\subsection{Explainability Methods}
There has been considerable amount of research on explaining the decisions of deep learning models.
For example, Ribeiro et al.~\cite{LIME} proposed LIME that can be applied to machine learning models in general, including deep learning models.
Furthermore, significant research on explainability methods that are designed for CNN models has been carried out.
For example, Simonyan et al.~\cite{saliency-maps} proposed Gradient-Based Saliency Maps.
This method simply differentiated the output of the model with respect to each pixel and created a heat map.
Another explainability method for CNN models is Grad-CAM~\cite{grad-cam}.
Grad-CAM used the values of feature maps in CNN models and the differential of the output with respect to them to calculate each pixel's importance. Then, a heat map was created. 

Contrary to the explainable models for CNNs, there are fewer works that explain GNN models.
For example, Pope et al.~\cite{node-cam} extended Grad-CAM to GNNs and calculated the importance of each node for the output of the GNN model.
Note that this approach is designed to explain the contribution of the nodes only.
GNNExplainer~\cite{GNNExplainer} is an explainability method for GNNs that explains which parts of edges and which dimensions of the node features are responsible for GNN model's outputs.
In addition, there are several approaches of explainability for edges in graphs.
Please refer to~\cite{yuan2020explainability} for a comprehensive survey of this field.

\section{PROPOSED METHODS}\label{sec:proposed}
In this section, we extend the explainability methods for CNNs to GNNs to predict which edges are important for GNN decisions.
We define an important edge as ``an edge that contributes to the increase of the GNN model's output.'' 

\subsection{LIME-Based Method}
First, we propose a LIME-based~\cite{LIME} explainability method for GNNs.
In the message passing operation described in the previous section, each node gathers the features of the adjacent nodes.
Thus, the edges are the paths through which node features pass.
Therefore, we define the operation to multiply node features passing through a certain edge by a weight $w \in [0, 1]$ as ``perturbing an edge.''
In the original LIME method, each part of the inputs is either removed completely or preserved.
The perturbing operation of multiplying information by continuous weight is different from the simple removal operation of the original LIME algorithm.

Let $n$ be the number of edges in the input graph $G$, $p$ be the probability of perturbing each edge, and $m$ be the number of samples.
$p$ and $m$ are both hyperparameters.
First, $m$ graphs in which each edge of $G$ is perturbed with the probability of $p$ are input to the GNN model.
Then, the combination of vectors $\bm{x}_k = [0, 1]^{n}$ that indicates which edge was perturbed and the output value of the model $y_k$ for $k = 1, 2, ..., m$ are obtained.
Here, each dimension of $\bm{x}_k$ corresponds to each edge of $G$ and shows the weight by which the information passing through each edge is multiplied.
A linear regression model is then constructed to predict $y_k$ from $\bm{x}_k$, and the importance of each edge is obtained as the coefficients of the linear regression model.
As the linear regression model, we use Lasso~\cite{Lasso}, which has a regularization term that limits the number of nonzero coefficients.
The loss function used for training the Lasso model $f$ is the weighted mean squared error (MSE):
\begin{equation}
\mathcal{L} =  \sum_{k=1}^{m} \exp \left(-\frac{(n - \|{\bf x}_k\|_1)^{2}}{\sigma^{2}}\right) \left(f\left({\bf x}_k\right) - y_k\right)^{2}
 + \lambda \|{\bf w}\|_1,
\end{equation}
where ${\bf w} \in \mathbb{R}^n$ represents $f$'s coefficients, and $\sigma$ and $\lambda$ are both hyperparameters.

\subsection{Saliency Map-Based Method}
In this section, we propose an extension of Saliency Maps~\cite{saliency-maps} to calculate each edge's importance.
Here, we consider GGNN~\cite{GGNN} as the model to be explained for example, and assume the model's output to be $y$. 
In GGNN, the message passing operation is represented as follows:
\begin{eqnarray}
\bm{h}_{v}^{t}={\rm GRU}\left(\bm{h}_{v}^{t-1}, \sum_{w \in N(v)} \bm{W} \bm{h}_{w}^{t-1}\right),
\end{eqnarray}
where $\bm{W}$ is a learnable matrix.
As the edges can be considered as pathways through which the node information propagates, one can assume that $\bm{W} \bm{h}_{w}^{t-1}$ passes through the edge $e_{v, w}$ connecting the node $v$ and the node $w$ at the $t$th message passing phase.
This operation is performed on all nodes in the graph, so the information that eventually passes through $e_{v, w}$ is $\bm{W} \bm{h}_{v}^{t-1}$ and $\bm{W} \bm{h}_{w}^{t-1}$.
Therefore, we consider the importance of $e_{v, w}$ to be the sum of information $\bm{W} \bm{h}_{v}^{t-1}$ and $\bm{W} \bm{h}_{w}^{t-1}$.
Let the size of $\bm{W}$ be $m$-by-$n$, the length of $\bm{h}_{v}^{t-1}$ be $n$, and the $k$th row component of $\bm{W} \bm{h}_{v}^{t-1}$ be ${\left(\bm{W} \bm{h}_{v}^{t-1}\right)}_k$.
Then, as the importance of $e_{v, w}$, $L_{edge} [t, v, w]$ is calculated as follows:
\begin{equation}
L_{edge} [t, v, w] = \sum_{k=1}^m \frac{\partial y}{\partial {\left(\bm{W} \bm{h}_{v}^{t-1}\right)}_k} 
+ \sum_{k=1}^m \frac{\partial y}{\partial {\left(\bm{W} \bm{h}_{w}^{t-1}\right)}_k}.
\end{equation}

\subsection{Grad-CAM-Based Method}
In this section, we propose an extension of Grad-CAM~\cite{grad-cam} to calculate each edge's importance.
Here, we consider the same GGNN~\cite{GGNN} model as in the previous subsection for example.
As in the Saliency Map-based method, the importance of $e_{v, w}$ can be considered as the sum of the importance of $\bm{W} \bm{h}_{v}^{t-1}$ and $\bm{W} \bm{h}_{w}^{t-1}$.
In this method, the importance of them is calculated by the method in~\cite{node-cam}, and then summed.
The specific method is described below:
$N$ feature vectors from $\bm{W} \bm{h}_{1}^{t-1}$ to $\bm{W} \bm{h}_{N}^{t-1}$ are converted to row vectors. Then, feature matrix $\bm{G}^t$ are created by stacking them.
The $(v, w)$ element of $\bm{G}^t$ corresponds to the $w$th row component of $\bm{W} \bm{h}_{v}^{t-1}$.
First, the weight $\alpha_{k}^{t}$ for $\bm{G}^t$'s column $k$ is calculated as follows:
\begin{equation}
\alpha_{k}^{t} = \frac{1}{N} \sum_{n=1}^{N} \frac{\partial y}{\partial \bm{G}_{n, k}^{t}}.
\end{equation}
Second, let $\bm{\alpha}^{t} \in \mathbb{R}^K$ be a vector whose $k$th row component is $\alpha_{k}^{t}$.
Then, the importance of $\bm{W} \bm{h}_{v}^{t-1}$ is obtained by calculating the dot product of $\bm{\alpha}^{t}$ and $\bm{W} \bm{h}_{v}^{t-1}$.
Finally, the importance of $e_{v, w}$, $L_{edge} [t, v, w]$ is calculated by adding the importance of vector $\bm{W} \bm{h}_{v}^{t-1}$ and $\bm{W} \bm{h}_{w}^{t-1}$ as follows:
\begin{equation}
L_{edge} [t, v, w] = \bm{\alpha}^{t} \cdot \bm{W} \bm{h}_{v}^{t-1} + \bm{\alpha}^{t} \cdot \bm{W} \bm{h}_{w}^{t-1}.
\end{equation}

\subsection{Baseline Methods}
We compare these methods to three baseline methods: GNNExplainer~\cite{GNNExplainer}, the removal method, and the random method.
In the removal method, edges are removed from the input graph one by one and are input to the model. Then, the importance of the edge is calculated by measuring the extent to which the output value decreases compared to the original value.
In the random method, each edge's importance is determined randomly.

\section{EXPERIMENTS}\label{sec:experiments}

\subsection{Experimental Setup}
We use three evaluation tasks: synthetic test, benzene ring test, and removal test.
In the synthetic test, we follow the setting in GNNExplainer~\cite{GNNExplainer} and use the dataset called BA-shapes.
This is a node classification dataset that contains a randomly generated graph with 300 nodes and 80 five-node ``house''-structured network motifs attached to it.
In this dataset each node has no node feature values and nodes in the base graph are labeled with 0; the ones located in the ``house'' are labeled with 1, 2, or 3.
First, a GCN~\cite{GCN} model is trained to predict each node's label.
If this model predicts that the node located in the ``house'' has a label other than 0, then the ground truth of the basis of this prediction can be regarded as the ``house''-structured motif.
The evaluation metrics is the percentage of the ground truth edges that are included in the top five edges in terms of importance calculated by each method (i.e., recall rate).

In the second evaluation task, the benzene ring test, we use the QM9 dataset~\cite{qm9} that contains molecule graphs with atoms as nodes and chemical bonds as edges.
We trained a GGNN~\cite{GGNN} model that performs binary classification if a molecule is aromatic.
As a molecule being aromatic is determined only by the presence of a benzene ring in the molecule, we can determine the ground truth of explanation as the five or six edges that form the benzene ring.
The evaluation metrics is the percentage of the ground truth edges that are included in the top five or six edges in terms of importance (i.e., recall rate).

The third evaluation task, the removal test, is motivated by the evaluation tasks of explainability methods for CNNs proposed by~\cite{np}.
First, the edges in the graph are removed from one to five in the order of importance of the edges obtained by each explainability method.
Second, these graphs are input to the GNN model to obtain the output values.
Subsequently, the number of removed edges is plotted on the horizontal axis, and the decrease in the predicted value compared to the original value is plotted on the vertical axis to obtain the Area Under the Curve ($\rm{AUC_{edge}}$).
The larger the $\rm{AUC_{edge}}$, the better the performance of the explainability method.
Let $y_k~(k = 0, 1, 2, ..., 5)$ be the output of the GNN model when $k$ edges are removed.
$\rm{AUC_{edge}}$ can be calculated by the following equation:
\begin{equation}
{\rm AUC_{edge}} = \sum_{k = 1}^{5} \frac{\left(y_0 - y_{k-1}\right) +\left(y_0 - y_{k}\right)}{2}.
\end{equation}
We use three datasets for the removal test: Cora~\cite{cora}, Coauthor~\cite{coauthor}, and Amazon~\cite{coauthor}.
Cora is a citation network dataset in which nodes are documents and edges are citation links.
Coauthor is a coauthorship network dataset in which nodes are authors and are connected by an edge if they coauthored a paper.
Amazon is a co-purchase graph dataset in which nodes are products and edges indicate that two goods are frequently bought together.
We trained three GCN~\cite{GCN} models that predict each node's label for these three datasets respectively.

\subsection{Results}
\begin{figure}[t]
 \begin{minipage}{0.32\hsize}
  \begin{center}
  \subfigure[LIME]{
   \includegraphics[width=27.5mm]{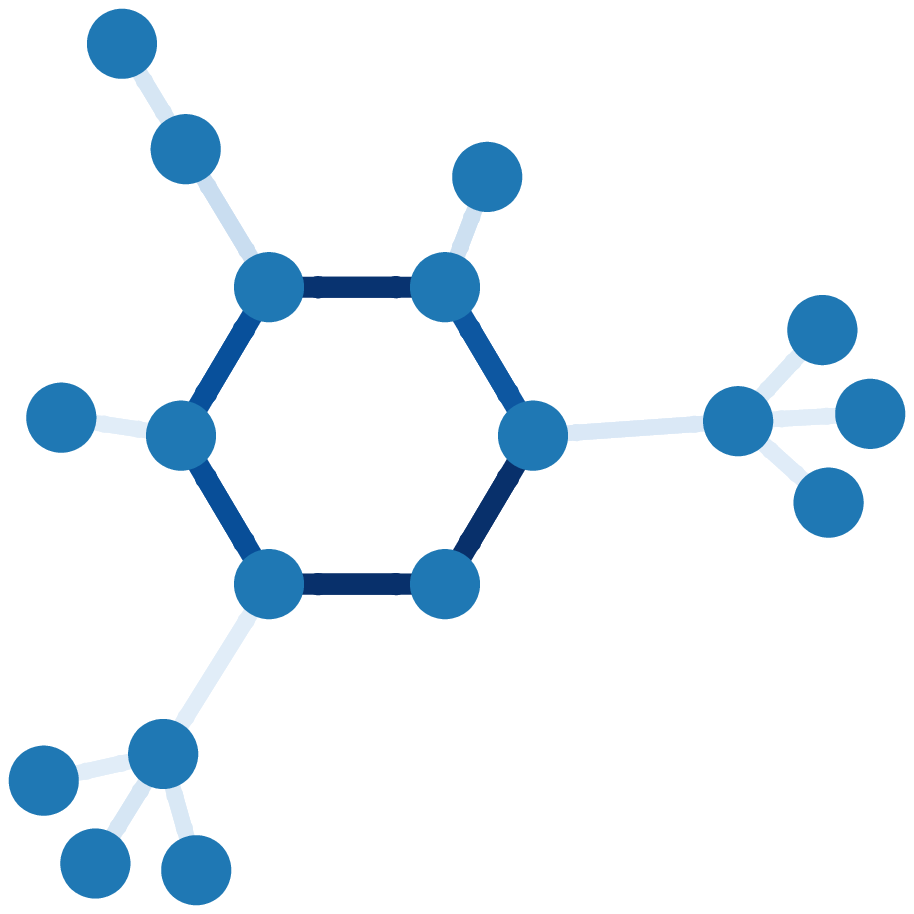}
   \label{fig:one}}
  \end{center}
 \end{minipage}
 \begin{minipage}{0.32\hsize}
  \begin{center}
  \subfigure[Grad-CAM]{
   \includegraphics[width=27.5mm]{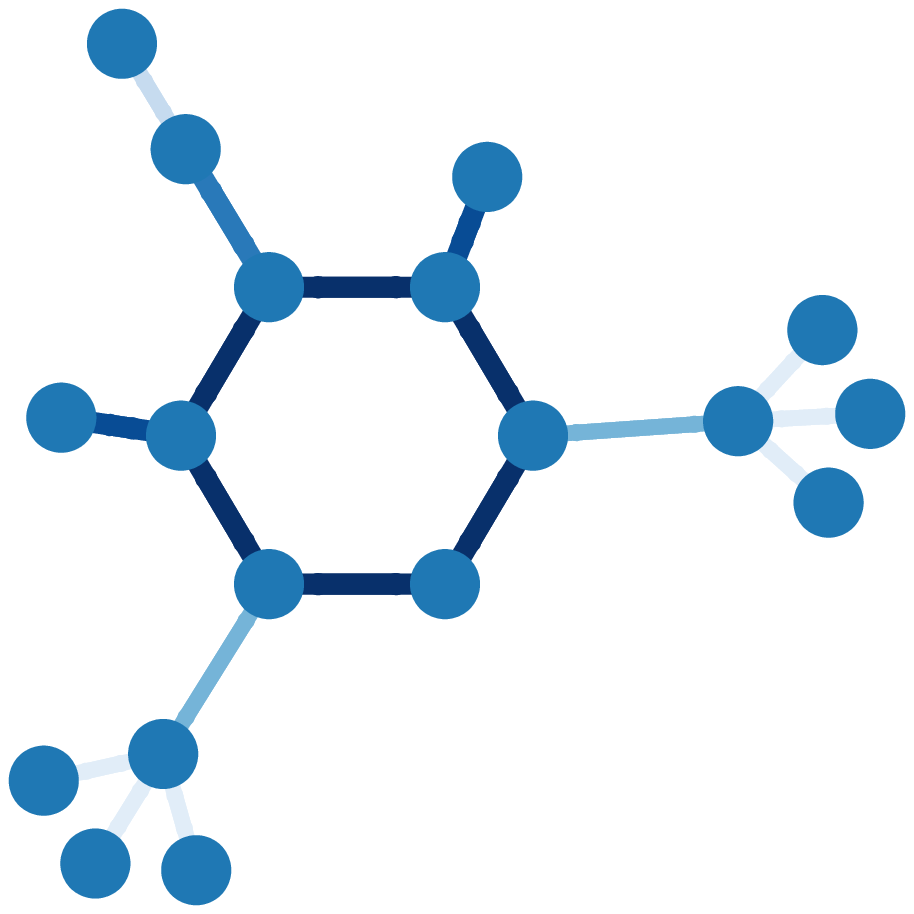}
   \label{fig:three}}
  \end{center}
 \end{minipage}
 \begin{minipage}{0.32\hsize}
  \begin{center}
  \subfigure[GNNExplainer]{
   \includegraphics[width=27.5mm]{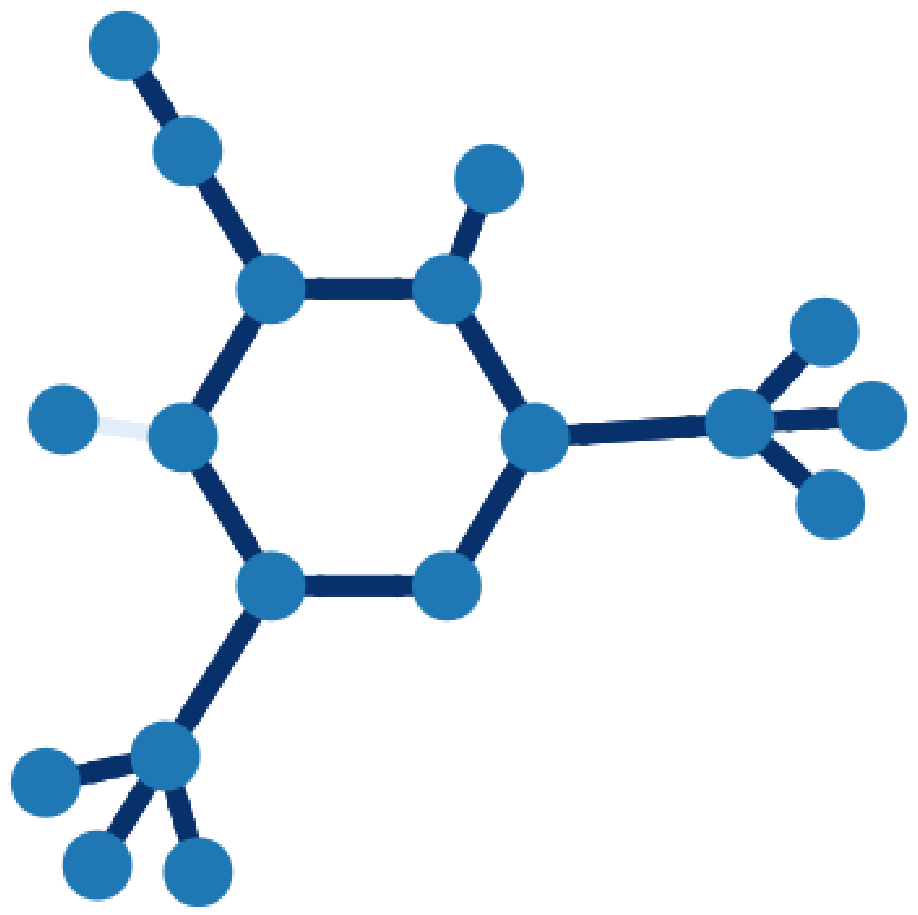}
   \label{fig:four}}
  \end{center}
 \end{minipage}
 \caption{Examples of explanation results for the benzene ring test.}
 \label{benzene-ex}
\end{figure}

Examples of explanation results for the benzene ring test are shown in Fig.~\ref{benzene-ex}, which shows that the importance of edges forming benzene rings is relatively high in all methods except for GNNExplainer.
In particular, LIME assigns high importance to the benzene ring edges only.
LIME can determine the important edges specifically.

{
\setlength\abovecaptionskip{-6pt}

\begin{table*}[t]
  \caption{Results of the synthetic test, benzene ring test, and removal test.}
  \label{scores}
  \begin{center}
  \begin{tabular}{cccccc}
    \toprule
     & Synthetic test (accuracy) & Benzene ring test (accuracy) & \multicolumn{3}{c}{Removal test ($\rm{AUC_{edge}}$)} \\
    \midrule
     Task & Node classification & Graph classification & \multicolumn{3}{c}{Node classification} \\
     Model & GCN~\cite{GCN} & GGNN~\cite{GGNN} & \multicolumn{3}{c}{GCN~\cite{GCN}} \\
     Dataset & BA-shapes~\cite{GNNExplainer} & QM9~\cite{qm9} & Cora~\cite{cora} & Coauthor~\cite{coauthor} & Amazon~\cite{coauthor} \\
    \midrule
    LIME~\cite{LIME} & 0.67 & $\bm{0.99}$ & $\bm{2.13}$ & $\bm{1.55}$ & $\bm{0.37}$\\
    Saliency Maps~\cite{saliency-maps} & 0.91 & 0.62 & 1.78 & 1.14 & 0.27\\
    Grad-CAM~\cite{grad-cam} & 0.20 & 0.88 & 0.21 & 0.03 & 0\\
    GNNExplainer~\cite{GNNExplainer} (Reproduction) & 0.87 & 0.44 & 0.57 & 0.60 & 0.08\\
    GNNExplainer~\cite{GNNExplainer} (Reported) & $\bm{0.93}$ & - & - & - & -\\
    Removal & 0.80 & 0.90 & $\bm{2.14}$ & $\bm{1.59}$ & $\bm{0.38}$\\
    Random & 0.15 & 0.36 & 0.18 & 0.02 & 0\\
  \bottomrule
\end{tabular}
\end{center}
\end{table*}

\begin{table}[t]
  \caption{The average computational costs of LIME and the removal method for each dataset.}
  \label{costs}
  \begin{center}
  \begin{tabular}{cccc}
    \toprule
    & Cora~\cite{cora} & Coauthor~\cite{coauthor} & Amazon~\cite{coauthor} \\
    \midrule
    LIME~\cite{LIME} & 0.98~s & 11.4~s & 79.3~s \\
    Removal & 3.63~s & 119.8~s & 704.6~s \\
    \bottomrule
\end{tabular}
\end{center}
\end{table}

}

The results of the synthetic test, benzene ring test, and removal test are shown in Table~\ref{scores}.
In the synthetic test, GNNExplainer (Reported) yields the best performance, followed by saliency maps.
However, in the benzene ring test, LIME shows the best performance followed by the removal method and Grad-CAM, whereas GNNExplainer is the worst among all methods except for the random method.
In the removal test, LIME and the removal method perform the best followed by saliency maps, whereas GNNExplainer and Grad-CAM are worse than these methods.
Although GNNExplainer performs well on the synthetic dataset, it does not perform well on real-world datasets. 
On the other hand, LIME is generally better than the other algorithms in our experiments.

In the removal test, the performances of LIME and the removal method are comparable, but the removal method requires high computational costs because it removes each edge one by one.
Table~\ref{costs} shows the average computational costs of LIME and the removal method in the three datasets for the removal test.
The computational cost of the removal method is about three to ten times larger than that of LIME.
Therefore, LIME is the best explainability method in terms of both performance and computational cost.

\subsection{Discussion}
LIME has the best performance among the three proposed methods in the real-world situations.
LIME directly perturb several edges in the input graph, therefore it can take interactions between edges into account.
In contrast, Grad-CAM and saliency maps calculate each edge's importance independently.
This capability of considering the interactions between edges would make the LIME's score the best.

However, in the synthetic test, GNNExplainer outperforms LIME.
In the synthetic dataset, each node has no feature values. This is different from the real-world datasets, where each node has unique feature values.
As mentioned in Section~\ref{sec:proposed}, edges are the paths through which node information passes, but no meaningful information passes through the edges in the synthetic dataset.
This absence of meaningful information passing through the edges would make the perturbing operation of LIME less effective.

\section{Conclusion}\label{sec:conclusion}
In this study, we extended explainability methods for CNNs to GNNs, i.e., LIME, Grad-CAM, and Gradient-Based Saliency Maps, to calculate each edge's importance for the outputs.
It was found that the performance of the LIME-based approach was the best in real-world situations in terms of both accuracy and computational cost. 

\renewcommand{\baselinestretch}{0.2}
\bibliographystyle{IEEEtran}
\bibliography{main}

\end{document}